\documentclass[letterpaper]{article} 
\usepackage{aaai23}  
\usepackage{times}  
\usepackage{helvet}  
\usepackage{courier}  
\usepackage[hyphens]{url}  
\usepackage{graphicx} 
\usepackage{natbib}  
\usepackage{caption} 
\usepackage{algorithm}
\usepackage{algorithmic}
\usepackage{xcolor, soul}
\sethlcolor{green}
\usepackage{subfigure}

\usepackage{newfloat}
\usepackage{listings}
\DeclareCaptionStyle{ruled}{labelfont=normalfont,labelsep=colon,strut=off} 
\floatstyle{ruled}
\newfloat{listing}{tb}{lst}{}
\floatname{listing}{Listing}
\title{NLP-based Decision Support System for Examination of Eligibility Criteria\\from Securities Prospectuses at the German Central Bank}
\author{
    Christian Hänig\textsuperscript{\rm 1},
    Markus Schlösser\textsuperscript{\rm 2},
    Serhii Hamotskyi\textsuperscript{\rm 1},
    Gent Zambaku\textsuperscript{\rm 2}, 
    Janek Blankenburg\textsuperscript{\rm 2}
}
\affiliations{

    \textsuperscript{\rm 1}Anhalt University of Applied Sciences, 
    \textsuperscript{\rm 2}Bundesbank\\
    christian.haenig@hs-anhalt.de, serhii.hamotskyi@hs-anhalt.de,\\
    markus.schloesser@bundesbank.de, gent.zambaku@bundesbank.de, janek.blankenburg@bundesbank.de
}

\begin{document}

\maketitle

\begin{abstract}
As part of its digitization initiative, the German Central Bank (Deutsche Bundesbank) wants to examine the extent to which Natural Language Processing (NLP) can be used to make independent decisions upon the eligibility criteria of securities prospectuses. Every month, the Directorate General Markets at the German Central Bank receives hundreds of scanned prospectuses in PDF format, which must be manually processed to decide upon their eligibility. 

We found that this tedious and time-consuming process can be (semi-)automated by employing modern NLP model architectures, which learn the linguistic feature representation in text to identify the present eligible and ineligible criteria. The proposed Decision Support System provides decisions of document-level eligibility criteria accompanied by human-understandable explanations of the decisions. The aim of this project is to model the described use case and to evaluate the extent to which current research results from the field of NLP can be applied to this problem. 

After creating a heterogeneous domain-specific dataset containing annotations of eligible and non-eligible mentions of relevant criteria, we were able to successfully build, train and deploy a semi-automatic decider model. This model is based on transformer-based language models and decision trees, which integrate the established rule-based parts of the decision processes. 

Results suggest that it is possible to efficiently model the problem and automate decision making to more than 90\% for many of the considered eligibility criteria. 
\end{abstract}

\section{Introduction}

Central banks play a crucial role in the global financial system by accepting various securities as collateral to implement monetary policy and manage liquidity. The prospectus of these securities provide details on the terms and conditions of issuance which are used to assess their eligibility regarding the Eurosystem eligibility criteria\footnote{https://eur-lex.europa.eu/legal-content/EN/TXT/PDF/ ?uri=CELEX:32014O0060\&qid=1664184198811\&from=EN}. Accurate determination of the eligibility of these securities is critical for both central banks and investors to make informed investment decisions. 

As part of its digitization initiative, the German Central Bank wanted to examine the extent to which this repetitive and time-consuming process can be automated. Besides the automation rate and accuracy of the decision, an important requirement is explainability of the decisions made. This allows domain experts to quickly evaluate (and, if required, easily correct) the decisions for the individual criteria. The resulting human feedback will then be fed back to the model training process as additional training data to incrementally improve the model's predictive quality.

This research paper aims to explore the automation of determination of the eligibility criteria of securities prospectuses of central banks. This paper will provide an in-depth analysis of the methodology used for estimating the eligibility, as well as an examination of current best practices and areas for improvement.

Every month, the Directorate General Markets of the German Central Bank receives hundreds of scanned prospectuses (in PDF format), which then must be manually processed to decide on their eligibility. In this project, we considered eight criteria - of varying complexity - to determine the eligibility of an emission. 

These criteria include:
\begin{itemize}
    \item Coupon
	\item Currency
    \item Early redemption amount
    \item Principal amount
    \item Redemption (amount) at maturity
    \item Special termination right
    \item Liquidation Status (Senior/Subordinated)
    \item Type of instrument
\end{itemize}

In this paper, we propose a Decision Support System that models the eligibility estimation process so that machine learning models can be integrated to automate decision making. For model training and evaluation, we created a human-annotated dataset for this use case. Additionally, we collected a large number of prospectuses to fine-tune a large language model (LLM) on German data from the financial domain. Finally, the Decision Support System is integrated into the business process for further evaluation.

\section{Data}
For our project, we created a dataset consisting of securities prospectuses from the financial domain mostly in German language. The dataset was created with the aim of creating a resource for training and evaluation of models that are required for automatic eligibility estimation of emissions. 

To the best of our knowledge, there is currently no annotated dataset that contains this kind of data.

\subsection{Data Collection}
The data collection process involved gathering securities prospectuses from various sources, including financial institutions and regulatory bodies. The prospectuses were collected from the websites of these institutions, as well as from publicly available databases. In total, we collected over 7000 prospectuses. All documents in the project were obtained in the form of PDF files. To enable further processing of the textual content, an optical character recognition (e.g. Adobe\footnote{https://www.adobe.com/}, Tesseract, see \citet{Smith2007:Tesseract}) component was utilized to extract the text from these files.

\subsection{Data Annotation}
Data annotation is an essential part of many AI projects. In our research, we focused on identifying relevant text passages for the aforementioned set of criteria and defined over 40 different annotation types that cover relevant text passages. We created comprehensive annotation guidelines that clearly defined each of these annotation types and provided examples. Throughout the annotation process, we continuously updated these guidelines to maintain a high level of consistency in the data. 

In total, more than 400 prospectuses were annotated. Only the pages containing relevant information were manually annotated, as many prospectuses consist of numerous pages. The remaining pages were discarded for model training and evaluation. For evaluation purposes, more than 50 prospectuses were annotated separately by two different human annotators, providing valuable insights into consistency and the complexity of the information extraction problem. Table \ref{table:dataset-stats} shows the final number of generated annotations. The number of test annotations results from annotating each test document twice. As merging annotations from different sources can be a challenging task, we decided to create four different test sets, each containing the annotations of a single human annotator.

\begin{table}[t]
\centering
\begin{tabular}{lrr}
Target type &  Train &  Test \\
\hline
coupon\_fixed & 431 & 375\\
coupon\_variable\_index & 56 & 84\\
coupon\_variable\_margin & 38 & 42\\
coupon\_variable\_operator & 37 & 43\\
coupon\_variable\_tenor & 45 & 75\\
currency & 514 & 577\\
early\_redemption\_amount & 64 & 52\\
early\_redemption & 177 & 108\\
isin & 421 & 417\\
principal\_amount & 784 & 800\\
redemption\_at\_maturity\_amount & 26 & 42\\
redemption\_at\_maturity & 370 & 347\\
special\_termination & 96 & 109\\
special\_termination\_amount & 61 & 63\\
status\_non\_preferred & 56 & 47\\
status\_senior\_non\_preferred & 488 & 333\\
type\_of\_instrument & 431 & 422\\
\end{tabular}
\caption{Number of manually generated annotations in the dataset. Each document in the test data was annotated by two different annotators.}
\label{table:dataset-stats}
\end{table}

\begin{figure}[ht!]
\centering
\includegraphics[width=1\columnwidth]{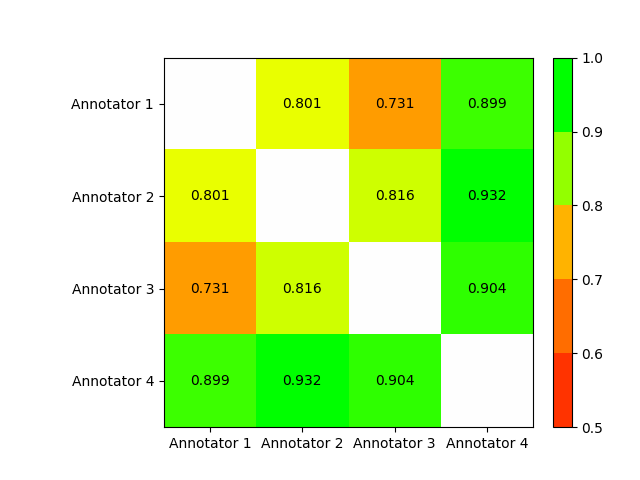} 
\caption{Average inter-annotator agreement scores.}
\label{fig:iaa-avg}
\end{figure}

To assess the consistency of the manual annotation process, we measured inter-annotator agreement. Inter-annotator agreement (IAA) is a metric for evaluating the reliability of manual annotation, as it provides insight into the degree of agreement among human annotators. In our case, we used Intersection over Union (IoU, see \citet{Braylan2022:IAA}). We choose IoU to better handle discontinuous annotations for which parts it is hard to define an order. IoU allows to compute the ratio of overlapping parts without having a deterministic and comparable order of the annotation parts by computing agreement based on bounding boxes of the different parts. The results for the most common annotation types are presented in Figure \ref{fig:iaa-avg}. As can be seen, the scores indicate a high level of agreement among the human annotators, with agreement scores between $0.731$ and $0.932$.

\begin{figure}[ht!]
     \centering
     \begin{subfigure}
         \centering
         \includegraphics[width=0.45\columnwidth]{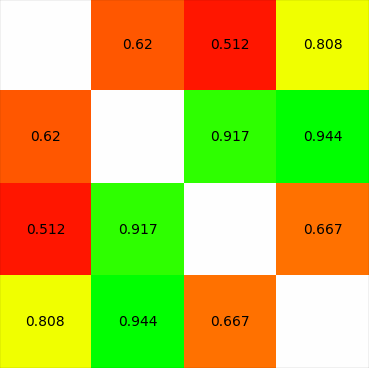}
     \end{subfigure}
     \begin{subfigure}
         \centering
         \includegraphics[width=0.45\columnwidth]{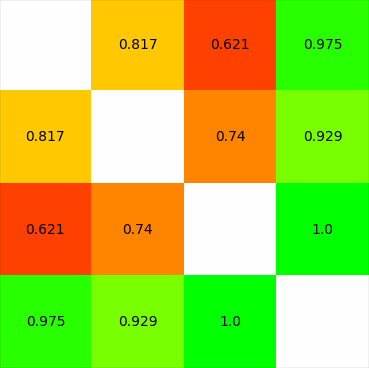}
     \end{subfigure}
\caption{Inter-annotator agreement scores for types \emph{redemption\_at\_maturity} (left) and \emph{status\_non\_preferred} (right).}
\label{fig:iaa-types}
\end{figure}

However, agreement scores suggest for some types (e.g. \emph{redemption\_at\_maturity} and \emph{status\_non\_preferred}, see Figure \ref{fig:iaa-types}) that the detection of relevant text passages is harder than for most other types. This observation can be explained by the annotators' different preferences about leading and/or trailing tokens around the most relevant tokens of an annotations. Additionally, the preferred location within the prospectus for checking particular criteria differs among the domain experts.

\subsection{Data Preprocessing}
Following the completion of the annotation process, a data extraction pipeline was developed to programmatically extract the JSON-formatted raw data containing the annotations from the annotation tool. The extracted data was then converted and transformed into a dataset for token classification. Specifically, the raw data was transformed into BIO-encoded sequences \cite{Ramshaw1995:BIO}, with the labels being aligned to the tokenization of the chosen transformer-based language model \cite{Vaswani2017:Transformer}. The dataset classes were implemented using the HuggingFace Datasets framework \cite{Lhoest2021:huggingfacedatasets} to facilitate efficient throughput during model training.

In the end, we discarded all annotation types with only few mentions as early experiments showed poor results for types with single-digits number of training instances.

\section{Model Architecture}
Our model architecture is divided into two parts: 

First, an ML model extracts text parts that are relevant for eligibility estimation. We call this part the \emph{Evidence Detection Model (EDM)}. 

Second, the \emph{Decided Model (DM)} processes the predictions of the Evidence Detection Model and makes a decision for each of the eight criteria. 

\subsection{Evidence Detection Model}
The evidence detection model is modelled as a Named Entity Recognition problem. We built our training/evaluation scripts using the HuggingFace library \cite{Wolf2020:huggingface}, and the different language models (LMs) we experimented with were, too, HuggingFace implementations. Lacking a German financial LM we experimented with a plethora of available language models to empirically determine the best model for this task. 

The following language models were included in our experiments and fine-tuned on our dataset:
\begin{itemize}
    \item FinBERT \cite{Liu2020:FinBERT}: a language model fine-tuned on English data from the financial domain, 
    \item BERT \cite{Devlin2019:BERT}: a general purpose English language model, 
    \item BERT-german and gbert \cite{Chan2020:GBERT}: two general purpose German language models.
\end{itemize}

\begin{figure}[ht]
\centering
\includegraphics[width=1\columnwidth]{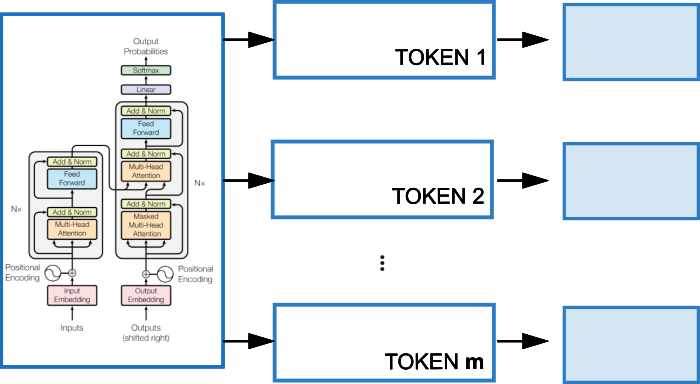} 
\caption{Binary Named Entity Recognition model. The classification layer projects representations of each of the $m$ tokens into the three \emph{BIO} labels for a single target type.}
\label{fig:model-binary}
\end{figure}

We followed the standard fine-tuning approach by using a pre-trained language model and adding a projection layer on top of it so that the representations are mapped into the three target classes of the BIO encoding. The resulting model structure is given in Figure \ref{fig:model-binary}. 

Our use case has additional complexities that precluded the use of an off-the-shelf solution. 

\begin{figure}[ht]
\centering
\includegraphics[width=1\columnwidth]{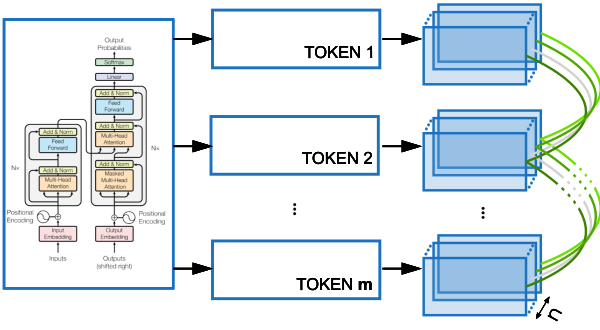} 
\caption{Multi-class multi-label model with consistent label predictions. The classification layer projects representations of each of the $m$ tokens into the three \emph{BIO} labels for each of $n$ target types. CRF layers provide consistent label transitions for all target types.}
\label{fig:model-future-work}
\end{figure}

A significant number of our annotations has overlaps, which created difficulties both at the annotation and at the model training stage. The source documents are PDFs that could contain tables or columns and transforming such complex layouts to plain-text exacerbated this problem. In standard sequence-labeling approaches, each token can belong to only one class. We mitigated this by training separate models for each annotation type (see Figure \ref{fig:model-binary}). In future work, we plan to use a multi-class multi-label model (see Figure \ref{fig:model-future-work}) that will predict labels separately for each of the \textit{n} classes. Additionally, a Conditional Random Field (CRF, \citet{Lafferty2001:CRF}) (depicted on the picture in green) can increase the consistency of predicted labels and thus, further increase the model's accuracy. 


\subsection{Decider Model}
The eligibility of the criteria provided in a prospectus is decided based on 8 criteria. For a prospectus to be eligible, all eight must be fulfilled. 

Six criteria can be directly validated if evidence of their eligibility is found in the prospectus. In these cases, the decider can make a decision based on the predictions of the Evidence Detection Model.\footnote{Example: if the EDM extracts \emph{EURO} as currency of the emission then this criterion is considered valid as EURO is a valid currency.} Only few instances of ineligible examples exist for these six criteria. If no evidence for these criteria is available in a document, then it gets marked for human evaluation.

The remaining two criteria are more complex. Along with the presence of annotations/predictions of multiple types (like the first six criteria), they include conditions such as issue date, issuer group, asset type etc. These two criteria are modeled through decision trees.

\section{Model Evaluation}
We evaluated different LMs and calculated precision, recall and F-score (PRF) for all annotation types. We used the calculated IAA scores (see Figures \ref{fig:iaa-avg} and \ref{fig:iaa-types}) as an upper margin of what we could realistically aim for. We measured the PRF scores for each of our test sets and computed their weighted average with respect to the corresponding test set size to obtain comparable scores representing the models' prediction quality.

\begin{table*}[ht]
\centering
\begin{tabular}{lrrrr}
Type &  bert-base-cased &  bert-base-german-cased &  finbert &  gbert-base \\
\hline
coupon\_fixed                    & 0.483 & 0.836 & 0.734 & \textbf{0.898} \\
coupon\_variable\_index          & 0.323 & 0.519 & 0.219 & \textbf{0.607} \\
coupon\_variable\_margin         & 0.327 & 0.634 & \textbf{0.647} & 0.561 \\
coupon\_variable\_operator       & 0.617 & 0.429 & 0.499 & \textbf{0.748} \\
coupon\_variable\_tenor          & 0.000 & \textbf{0.774} & 0.596 & 0.770 \\
currency                         & 0.896 & 0.931 & \textbf{0.954} & 0.942 \\
early\_redemption                & 0.535 & 0.648 & 0.547 & \textbf{0.769} \\
early\_redemption\_amount        & 0.181 & 0.431 & 0.000 & \textbf{0.554} \\
isin                             & 0.883 & 0.877 & 0.868 & \textbf{0.927} \\
principal\_amount                & 0.833 & 0.921 & 0.916 & \textbf{0.924} \\
redemption\_at\_maturity         & 0.566 & 0.765 & 0.531 & \textbf{0.775} \\
redemption\_at\_maturity\_amount & 0.000 & 0.746 & 0.000 & \textbf{0.761} \\
special\_termination             & 0.683 & \textbf{0.712} & 0.628 & 0.665 \\
special\_termination\_amount     & 0.520 & \textbf{0.813} & 0.679 & 0.680 \\
status\_non\_preferred           & 0.222 & \textbf{0.633} & 0.556 & 0.438 \\
status\_senior\_non\_preferred   & 0.718 & 0.822 & 0.782 & \textbf{0.846} \\
type\_of\_instrument             & 0.752 & 0.800 & 0.726 & \textbf{0.821} \\
\hline
mean                             & 0.502 & 0.723 & 0.581 & \textbf{0.746} \\
\end{tabular}

\caption{Comparison of evaluation results achieved by model fine-tuning with respect to the employed language models. Scores are averaged F1-scores weighted by number of instances of the target type. The bottom rows contain macro-averaged F1-scores per language model.}
\label{table:eval-prf}
\end{table*}

Our experimental findings (see Table \ref{table:eval-prf}) can be summarized as follows:
\begin{itemize}
    \item The embeddings of FinBERT seem to encode financial properties better than a general English language model (BERT).
    \item Both general purpose German language models outperform the English language models.
\end{itemize}

The comparison of achieved F-Scores with inter-annotator agreement scores revealed that 11 of 17 models produced comparable results. Given the better performance of FinBERT compared to BERT, we hypothesize that utilizing a German language model specific to the financial domain could lead to further improvement in model performance.

\section{Deployment}
We serve the model as two REST API endpoints using FastAPI\footnote{https://fastapi.tiangolo.com} in the backend. The first takes a raw document without any annotations and provides the model’s predictions as JSON-formatted response. The second expects a document that already contains annotations, and the model only performs the decision part (DM) based on the given annotations.
The response object, used by both endpoints, contains the final verdict (eligible or not eligible) for the full prospectus, and the decision together with further explanation for every single criterion. The model can be deployed containerized and by the REST interface included as a standalone service.

\section{Integration and Frontend}
Our aim is to optimize the current process by providing the experts, who currently perform the manual review, with a tool that supports them in their daily work and thus reduces the workload while maintaining/improving the quality of the reviews. For this purpose, we are developing a tool that automatically processes and manages securities prospectuses, integrates the trained model, and visualizes the results to the user. The focus is on making the decision, predicted by the model, comprehensible and thus verifiable for the users. 

For this purpose, we build a web-frontend consisting of three levels: 

\textbf{1. The document-level decision is presented as a boolean value (eligible or ineligible).} 

\textbf{2. The decision for each individual criterion} structured according to whether the criterion argues for or against the eligibility. For each criterion, the user is shown which value has been identified by the model (e.g., "Euro" as currency) as well as a confidence score. In addition, all locations where the model has located a corresponding entity are displayed to the user (including the confidence score). This means that there are also locations listed where the model has identified a different value for a criterion as the value with the highest confidence score. This makes the user aware that there are other possible values for the criterion in the prospectus besides the recognized value and thus gives the user the opportunity to easily verify that the correct value for the criterion has been selected from the list of values found. 

\textbf{3. The original PDF is displayed.} All the entities localized by the model are highlighted and labeled, so that it can be easily checked whether the correct value was recognized for each criterion, where more context can be taken into account than is possible in the 2nd explanation level. In addition, the user has the possibility to confirm, to edit or to add new annotations. Based on the manually revised annotations, the decision trees are triggered again (using the second endpoint of the model API) and the decision of the model for the individual criteria as well as for the overall result is updated.

\begin{figure}[t]
\centering
\includegraphics[width=0.9\columnwidth]{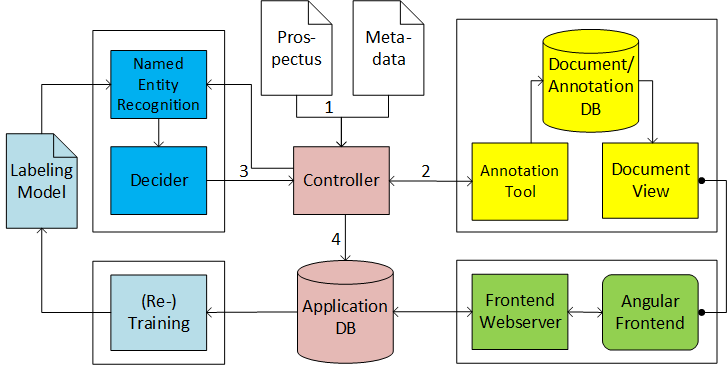}
\caption{Architecture of the overall system}
\label{M52_Architecture}
\end{figure}

The frontend and the model are embedded in a Decision Support System, that consists of three strongly decoupled subsystems:

\textbf{1. Processing}: Here, new securities prospectuses together with external meta data are gathered (see Figure \ref{M52_Architecture}, Step 1), analyzed and the results are made available in the Application Database. There we developed one central element (the so called controller) serving the different components (model, annotation tool, database), so that the components themselves, which are integrated as services, are completely decoupled.\footnote{Loosely following the mediator pattern \cite{Gamma1995}} In the 2nd step, the new data is added to the Document Database via the annotation tool and returned to the controller as a JSON object. This is passed to the model service, which executes the decision process as described above. The response is passed to the controller, where it is enriched with the metadata and saved in the Application Database (see Steps 3 and 4).

\textbf{2. Frontend}: We are using a standard web application stack to display the data from the Application Database via a web server in an Angular\footnote{https://angular.io/} web-frontend. For displaying prospects including the recognized annotation we are using a document view provided by the annotation tool. This can also be used to manually adjust (or confirm) the annotations.

\textbf{3. Retraining}: The third component extracts new documents from the Application Database for which experts from the Directorate General Markets have adapted or confirmed the model results, so that it can be assumed that the annotations in the database are correct. With these documents the training data is extended and in the future a retraining will be automatically executed on the extended data. This results in an updated model artifact that will be integrated into the model service via a CI/CD pipeline.

Each of these three components is developed and containerized-deployed separately so that every component is fully functional without the two others. More precisely, already processed data can be displayed and updated without the processing subsystem (or any of its components). The only component required by all three subsystems is an available Application Database.

\section{Challenges and Limitations}
It is worth mentioning that we went through some challenges throughout the design of the Decision Support System. 

First, there is no ground truth dataset that we could use to train our models upon. Thus, a time-consuming annotation process was conducted by domain experts of the Directorate General Markets at the German Central Bank, resulting in a rich dataset composed of thousands of annotations. 

Second, we realized that the annotations were unbalanced. We found out there were quite a lot of annotations for certain criteria, but very few for other criteria. We tackled this imbalance by additionally annotating documents containing rare mentions of criteria, thus, expanding and enriching the dataset. Although this improved noticeably the overall model accuracy, some criteria still do not have the desired number of annotations to train on. 

Third, as the documents are present in heterogeneous formats including structural elements like tables, check boxes, etc., the textual flow cannot be extracted in a consistent way for all documents. This states a challenge for document understanding in general. We plan to integrate dedicated approaches for structure analyses of documents, e.g., table detection. Finally, the annotation tool that we used does not support overlapping annotations, which introduced additional effort to data management of multiple versions of the same document with different annotations.

\section{Conclusion and Future Work}

In conclusion, our project has demonstrated that fine-tuning existing out-of-domain language models exhibit superior performance in comparison to in-domain language models of a different language. Our findings indicate that the manual determination of eligibility criteria of securities prospectuses can be assisted by machine learning models.

The integration of the proposed decision support system into the business process is expected to provide ample opportunities for improvement through the provision of additional training data and human feedback. 

Additionally, we want to emphasize the importance of proper document structure modelling in document understanding tasks, which calls for further investigation. 

Furthermore, we want to investigate the potential gains of multi-class multi-label models with CRFs to enhance the overall prediction quality of our models. Finally, creating a German in-domain language model for financial use cases seems particularly important for advancing the state-of-the-art in this application field.

\bibliography{aaai23}

\end{document}